\pdfoutput=1

\documentclass[11pt]{article}

\usepackage{acl}

\usepackage{times}
\usepackage{latexsym}
\usepackage{multirow}
\usepackage{booktabs}

\usepackage{graphicx}

\usepackage{enumitem}

\usepackage[T1]{fontenc}

\usepackage[utf8]{inputenc}

\usepackage{microtype}

\usepackage{color}
\usepackage{algorithm}
\usepackage[noend]{algpseudocode}
\usepackage[utf8]{inputenc} 
\usepackage[T1]{fontenc}    
\usepackage{hyperref}       
\usepackage{url}            
\usepackage{booktabs}       
\usepackage{amsfonts}       
\usepackage{nicefrac}       
\usepackage{microtype}      
\usepackage[linewidth=1pt]{mdframed}
\usepackage{tikz-dependency}
\usepackage{amsmath,amsfonts,bm}
\usepackage{graphicx}
\usepackage{algorithmicx}
\usepackage[noend]{algpseudocode}
\usepackage{booktabs}
\usepackage[percent]{overpic}
\usepackage{subcaption}
\usepackage{siunitx}
\usepackage{wrapfig}
\usepackage{listings}

\definecolor{RoseQuartzBg}{HTML}{F7CAC9}
\definecolor{RoseQuartz}{HTML}{F5A798}
\definecolor{Serenity}{HTML}{92A8D1}
\definecolor{OrangeRed}{rgb}{1.0, 0.27, 0.0}
\definecolor{Red}{rgb}{1.0, 0.0, 0.0}
\definecolor{Turquoise}{HTML}{0F4C81}
\usepackage{xparse}
\NewDocumentCommand{\mo}{ mO{} }{\textcolor{black}{\textsuperscript{\textit{Mo}}\textsf{\textbf{\small[#1]}}}}

\NewDocumentCommand{\yao}{ mO{} }{\textcolor{red}{\textsuperscript{\textit{Yao}}\textsf{\textbf{\small[#1]}}}}

\NewDocumentCommand{\gy}{ mO{} }{\textcolor{purple}{\textsuperscript{\textit{GY}}\textsf{\textbf{\small[#1]}}}}

\NewDocumentCommand{\gc}{ mO{} }{\textcolor{purple}{\textsuperscript{\textit{Chuang}}\textsf{\textbf{\small[#1]}}}}

\newcommand{\modelname}{DRRN-\textsc{LoG }}
\newcommand{\modelnamens}{DRRN-\textsc{LoG}}

\title{Revisiting the Roles of ``Text'' in Text Games}

\author{Yi Gu\thanks{\quad Equal contribution.}$\,\,^{1}$\quad Shunyu Yao$^{*2}$\quad Chuang Gan$^{3,4}$\quad Joshua B. Tenenbaum$^{5}$\quad Mo Yu$^{6}$\\
$^1$UC San Diego \quad
$^2$Princeton University 
$^3$MIT-IBM Watson AI Lab \\
  \quad $^4$UMass Amherst \quad $^5$MIT \quad $^6$WeChat AI \\
\texttt{yig025@ucsd.edu,shunyuy@princeton.edu,moyumyu@tencent.com}
}

\begin{document}
\maketitle
\begin{abstract}
Text games present opportunities for natural language understanding (NLU) methods to tackle reinforcement learning (RL) challenges. However, recent work has questioned the necessity of NLU by showing random text hashes could perform decently. In this paper, we pursue a fine-grained investigation into the roles of text in the face of different RL challenges, and reconcile that semantic and non-semantic language representations could be complementary rather than contrasting. Concretely, we propose a simple scheme to extract relevant contextual information into an approximate state hash as extra input for an RNN-based text agent. Such a lightweight plug-in achieves competitive performance with state-of-the-art text agents using advanced NLU techniques such as knowledge graph and passage retrieval, suggesting non-NLU methods might suffice to tackle the challenge of \emph{partial observability}. However, if we remove RNN encoders and use approximate or even ground-truth state hash alone, the model performs miserably, which confirms the importance of semantic function approximation to tackle the challenge of \emph{combinatorially large 
observation and action spaces}. Our findings and analysis provide new insights for designing better text game task setups and agents.
\end{abstract}

\section{Introduction}
\label{sec:intro}
\begin{figure}[t]
  \centering
  \includegraphics[width=0.49\textwidth]{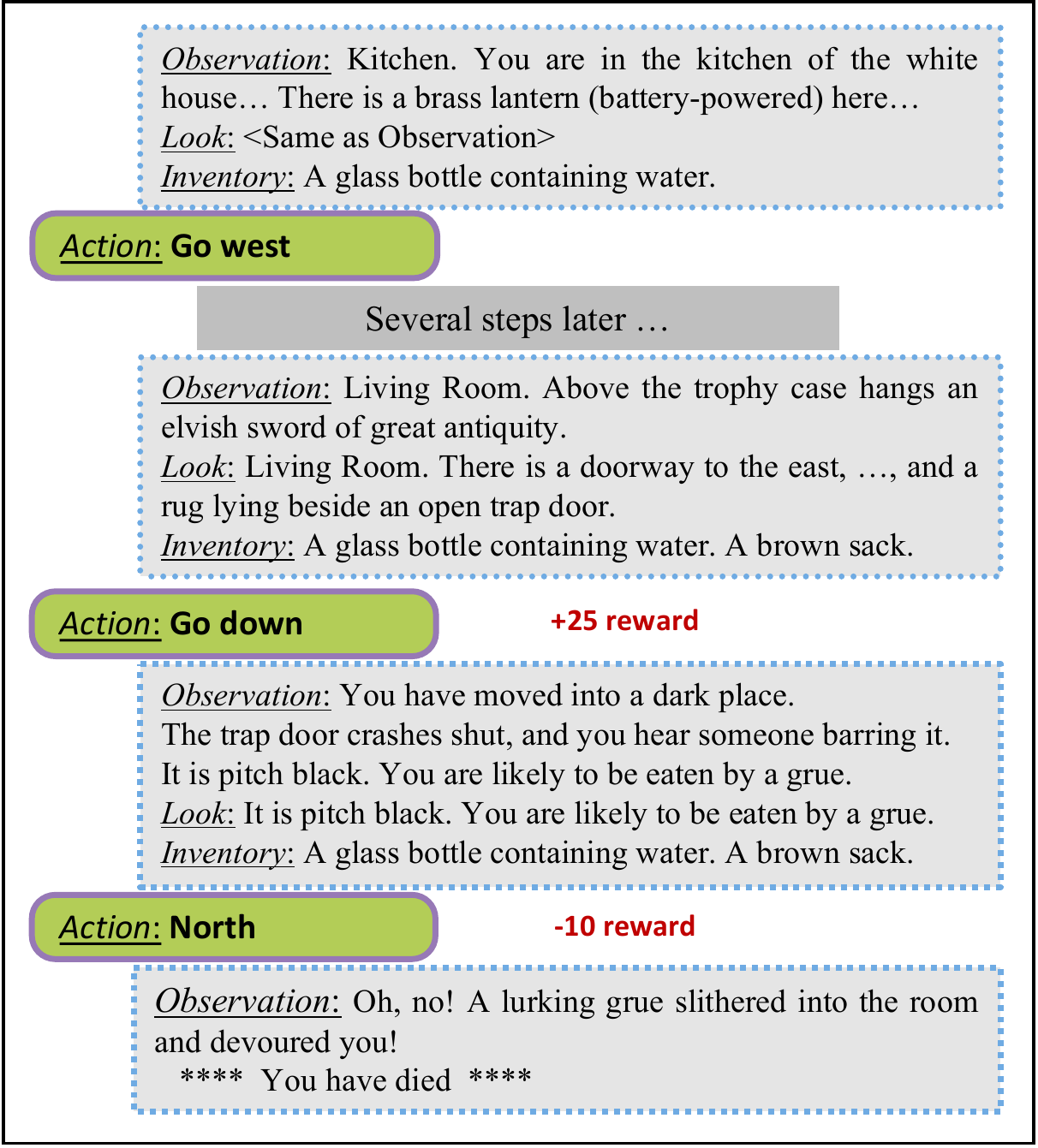}
    \vspace{-3mm}
  \caption{A game trajectory from Zork I.} 
  \label{fig:game}
  \vspace{-6mm}
\end{figure}

\textcolor{black}{In text-based games~\cite{narasimhan2015language,he2016deep,hausknecht2019interactive,cote2018textworld}, players read text observation, command text actions to interact with a simulated world, and gain rewards as they progress through the story. 
From a reinforcement learning (RL) viewpoint, they are partially observable Markov decision processes (POMDP)
--- the current observation does not carry the full information of the game progress. 
In our Figure~\ref{fig:game}  example, visiting the \texttt{Living Room} before or after the \texttt{dark place} puzzle may yield the same observation, but only when informed by the game history, the player can decide whether \texttt{go down} is the right action.}


Recent work has proposed to incorporate RL agents with natural language understanding (NLU) capabilities for better text game performance.
For example, pre-trained language models support \emph{combinatorial action generation}~\cite{yao2020keep}; commonsense reasoning~\cite{murugesan2021efficient}, information extraction~\cite{ammanabrolu2020graph}, and reading comprehension~\cite{guo2020interactive} provide priors for exploration with \emph{sparse reward} and \emph{long horizon}; and knowledge graph~\cite{ammanabrolu2020graph} and document retrieval~\cite{guo2020interactive} techniques help alleviate \emph{partial observability}. 

Nevertheless, \citet{yao2021reading} doubts the need of NLU for RL agents trained and evaluated on the same game. They found that a text game agent, DRRN~\cite{he2016deep}, performs even slightly better when RNN-based language representations are replaced with non-semantic hash codes.
Intuitively, hash serves to \emph{memorize} state-action pairs and ignore text similarities, which is sometimes useful --- consider the second-to-last observation in Figure~\ref{fig:game} and a counterfactual observation where a ``\texttt{lantern}'' is added into ``\texttt{Inventory}'', RNNs might encode them very similarly though they lead to antipodal consequences (die or explore the underground). How do we reconcile this with recent NLU-augmented text agents with improved performances? Where are semantic representations useful, and where would a hash approach suffice?

In this paper, we present initial findings that semantic and non-semantic language representations could work hand-in-hand better than each alone by targeting different RL challenges.
Concretely, we show the hash idea could help DRRN tackle \emph{partial observability} -- returning to the Figure~\ref{fig:game} example, to get lantern to avoid death, it is vital to know where the lantern is, which is revealed in a previous instead of current observation.
Based on such intuition, we propose a simple algorithm that tracks the current location and the up-to-date descriptions of all locations, then encode them into a single approximate state hash vector as extra DRRN input.
Though lightweight and easy-to-implement, such a representation plug-in improves DRRN scores by 29\% across games, with competitive performances against state-of-the-art text agents using advanced NLU techniques and pre-trained Transformer models. The effectiveness is further confirmed by comparing to models that plug in groundtruth state or location hash codes, where we find our performance and these upper bounds with very little gaps. These results suggest that the current \emph{partial observability} bottlenecks might not require advanced NLU models or semantic representations to conquer. 


However, such a message is gauged by the ablations that show the approximate state hash alone only achieves 58\% of the full performance, as it fails to handle other RL challenges such as \emph{the combinatorial state and action spaces}. 
In conclusion, we find the role of NLU in text games is not black-or-white as indicated by prior work, but rather differs for different RL challenges, and agents could benefit from combining semantic and non-semantic language representations that target different functionalities. Our results and insights contribute to future research in designing better tasks and models toward autonomous agents with grounded language abilities.



\section{Preliminaries}

\subsection{Problem Formulation}
\label{ssec:formulation}
A text game can be formulated as a partially observable Markov decision process (POMDP) $\langle S, A, T, O, \Omega, R, \gamma \rangle$, where at the $t$-th turn the agent reads a textual observation $o_{t} = \Omega(s_t) \in O$ as a partial reflection of underlying world state $s_t \in S$, issues a textual command $a_{t} \in A$ in response, and receives a sparse scalar reward $r_{t} = R(s_t, a_t)$ in light of game progress. The state transition $s_{t+1} = T(s_t, a_t)$ is hidden to the agent. The goal is to maximize the expected cumulative discounted rewards $\mathbf{E}[ \sum_{t} \gamma^{t} r_{t}]$.

\paragraph{Observations and States} 
Following prior practice in the Jericho benchmark~\cite{hausknecht2019interactive}, we augment the direct observation $o_{t}$ with inventory $i_t$ and location description $l_t$ obtained by issuing actions ``\emph{inventory}'' and ``\emph{look}'' respectively. But even this may not reveal the complete $s_t$ (Section~\ref{sec:intro}), which is Jericho 
includes an object tree and a large simulator RAM array hidden to players. As $s_t$ is large and lacks interpretability, more often used is the state hash $\mathrm{h}(s_t)$, where $\mathrm{h}: S \to \mathbb{N}$ maps each state to an integer that can be used to probe if states are identical, but cannot provide \emph{semantic} information about state differences. Access to $s_t$ or $\mathrm{h}(s_t)$ is a handicap in Jericho.

\subsection{The DRRN Baseline and its Hash Variant}
\label{ssec:drrn}
Denote $c_t = (o_1, a_1, \cdots, o_t)$ as the game context up to $o_t$, and for convenience we omit the subscript $t$ when no confusion is caused.
Our baseline RL model, Deep Reinforcement Relevance Network (DRRN)~\cite{he2016deep}, learns a Q-network
\begin{equation}
    Q(c_t, a_t) = \mathrm{MLP}(sr, ar)
\vspace{-3mm}
\end{equation}
where the \textbf{state and action representations} 
\begin{equation}
\begin{aligned}
\label{eq:sr_drrn}
    sr_{\text{\small drrn}} 
    &
    = [\mathrm{GRU}_{1}(o_t), \mathrm{GRU}_{2}(i_t), \mathrm{GRU}_{3}(l_t)] \\
    ar_{\text{\small drrn}} 
    &
    = \mathrm{GRU}_4(a_t)  
\vspace{-3mm}
\end{aligned}
\end{equation}
are encoded by gated recurrent units (GRU)~\cite{cho2014properties}. The temporal difference (TD) loss and Boltzmann exploration are used for RL.
In \citet{yao2021reading}, \textcolor{black}{Eq. \ref{eq:sr_drrn}} is replaced by random, fixed, non-semantic hash representations
\begin{equation}
\begin{aligned}
    \label{eq:sr_hash}
    sr_{\text{\small hash}} &= [H(o_t), H(i_t), H(l_t)] \\
    ar_{\text{\small hash}} &= H(a_t)
    \vspace{-3mm}
\end{aligned}
\end{equation}
where a hash vector function $H = vec \circ h$ first maps inputs to integers (via Python built-in hash) then to random normal vectors (by using the integer as the generator seed). 
However, neither of the models addresses  partial observability by using the context $c_t$ beyond the current observation $o_t$.

\begin{table*}
\centering
\small
\resizebox{0.9\textwidth}{!}{
\begin{tabular}{l|cccccccc||cccc||cc}
\toprule
\multirow{3}{*}{\textbf{Game}} &   \multicolumn{8}{c||}{\bf \emph{DRRN and Variants}}  & \multicolumn{4}{c||}{\bf \emph{Agents with advanced NLU}} & \multirow{3}{*}{\textbf{Max}} \\
& \multicolumn{2}{c}{\textbf{DRRN}}& \multicolumn{2}{c}{\textbf{Obs Hash}}& \multicolumn{2}{c}{\textbf{+ Inv-Dy}} & \multicolumn{2}{c||}{\bf LoG (ours)} & \multicolumn{2}{c}{\textbf{MPRC-DQN}} & \multicolumn{2}{c||}{\textbf{KG-A2C}} & \\
&\bf Avg & \bf Max&\bf Avg & \bf Max&\bf Avg & \bf Max&\bf Avg & \bf Max&\bf Avg & \bf Max&\bf Avg & \bf Max\\
\midrule
zork1  & 39.4&53 & 35.5&50 & 43.1&87 & \textbf{51.2}&\textbf{107} & 38.3&-- & 33.6&35 & 350 \\
zork3  & 0.4&4.5 & 0.4&4 & 0.4&4 & {1.33}&\textbf{5} &  \textbf{3.0}&\textbf{5.0} & 0.1&-- & 7 \\
pentari  & 26.5&45 & \textbf{51.9}&\textbf{60} & 37.2&50 & 44.4&\textbf{60} &44.4&-- & {48.2}&56 & 70 \\ 
detective  & 290&\textbf{337} & 290&317 & 290&323 & 288.8&313.3 & \textbf{317.7}&-- & 246.1&274 & 360\\
ludicorp  & 12.7&23 & 14.8&23 & 13.5&23 & {16.7}&23 & 10.9&\textbf{40.7} & \textbf{17.6}&19 & 150\\ 
inhumane & 21.1&45 & 21.9&45 & 19.6&45 & 25.7&\textbf{56.7} & \textbf{29.8}&53.3 & 3&-- & 90 \\
\midrule
avg norm & .28&.52 & .34&.52 & .30&.51 & \bf{.36}&.\bf{59} & .41&- & .27&- \\
\bottomrule
\end{tabular}
}
  \vspace{-2mm}
\caption{\small{Final episodic/maximum explored scores for different games. MPRC-DQN numbers with max scores correspond to version change of games, so we re-run their model and report the new results. Average normalized score (avg norm) is model score divided by maximum game score, averaged across games.}}
\label{tab:overall}
  \vspace{-4mm}
\end{table*}

\section{Method}

The key to handle partial observability is to extract the appropriate state-distinguishing information from the context $c_t$ --- while under-extraction leads to different states with same representations, over-extraction leads to diverging representations for the same state with different history paths. So to approximate the state hash, we first obtain and maintain a location map by exploration with limited depth $d$, collecting names of adjacent rooms:       \vspace{-1.5mm}
\begin{equation}
    po_1 = \{ (p, loc(c_t, p) \mid p \subset A^{d} \}
    \label{eq:po1}
      \vspace{-1.5mm}
\end{equation}
where $p$ is a sequence of navigation actions, and $loc$ is the location after following $p$ from $c_t$. Essentially, $po_1(c_t)$ serves to distinguish different locations with same names.\footnote{\textcolor{black}{For a common example, the game designers could set mazes with same room names.}}

Secondly, we collect the most-recent location descriptions for all locations, so that we may know, for example, the whereabouts of the lantern when needed (Section~\ref{sec:intro}). \vspace{-1.5mm}
\begin{equation}
    po_2 = \{ (loc, \text{LastLook}(loc)) \mid loc \in \text{Map} \} 
    \label{eq:po2}
      \vspace{-1.5mm}
\end{equation}

Together, our model \textcolor{black}{DRRN-\underline{Lo}cation\underline{G}raph (\textsc{LoG})} takes state representation \vspace{-1.5mm}
\begin{equation}
    sr_{\text{WH}} = [sr_{\text{drrn}}, {H}(po_1)), {H}(po_2)] 
\vspace{-1.5mm}
\end{equation}

The algorithm details are in Appendix~\ref{app:algorithm}.

\section{Experiments}
\label{sec:exp}

\paragraph{Implementation Details}
We adopt DRRN hyperparameters from \citet{yao2021reading} to train our model.
Following previous work, we implement the BiDAF~\cite{seo2016bidirectional} attention mechanism
and the inverse dynamics auxiliary objective~\cite{yao2021reading} for better text encoding.
The episodic limit is 100 steps and the training has 1,000 episodes from 8 parallel game environments. 
For $po_1$, we use $d=1$ as depth limit.
We train three independent runs for each game.
More details are in {Appendix}~\ref{app:detail}.



\paragraph{Baselines}
Our approach builds on the backbone DRRN agent, thus we provide \textcolor{black}{fair} comparisons to the original DRRN and its hash and inverse dynamics variants from \citet{yao2021reading}.
We also compare with more complex state-of-the-art agents that are designed to deal with the partial observability via NLU:
\begin{itemize}[leftmargin=*]
\setlength\itemsep{0em}
    \item \textbf{MPRC-DQN}~\cite{guo2020interactive}, which retrieves the relevant history to enhance the current observation, and formulates the action prediction as a multi-passage reading comprehension problem. 
    \item \textbf{KG-A2C}~\cite{ammanabrolu2020graph,ammanabrolu2020avoid}, which extracts an object graph with OpenIE~\cite{angeli2015leveraging} or a BERT-based QA model~\cite{devlin2018bert}, and embeds the graph to a single vector as the state representation.
    We compare with the better result from the two papers for each game.
\end{itemize}

\paragraph{Evaluating Games}
We select 6 games from Jericho \cite{hausknecht2019interactive} where \textbf{MPRC-DQN} or \textbf{KG-A2C} exhibits performance boosts, thus are more likely to suffer from partial observability.

\subsection{Game Results}
Table~\ref{tab:overall} shows game scores for all models. Among DRRN and it variants, \modelname performs best on {4 of the 6 games}. 
%
More impressively, our agent is competitive against MPRC-DQN (better or same score on 3/6 games) and KG-A2C (better scores on 4/6 games) in terms of winning rates. 
Overall, our \modelname achieves the second best average normalized score of $36\%$, only behind $41\%$ of MPRC-DQN (which is largely attributed to Zork III).
Considering the fact that we explicitly choose the six games in favor of these two state-of-the-art baselines, such a result indicates that advanced NLU techniques might not be a must to solve partial observability --- at least in the scoring ranges of current text game agents (i.e.\,average normalized score less then $50\%$).

%
%




\subsection{Oracle Analysis with Groundtruth States}
\label{ssec:upperbound}
\textcolor{black}{Next, we aim to study the performance gap between our model, and an oracle version that replaces our approximate state hash with the groundtruth state hash (GT-State) from Jericho.}
The GT-States could perfectly distinguish different states apart, where $sr_{gt} = (sr_{\text{drrn}}, H(s)).$

As shown in Table~\ref{tab:gt}, the scores of \modelname and the GT-State are very close across different games, meaning our approximation has been close-to-perfect within the state hashing scheme.
Notably, even GT-State fails to totally surpass \textbf{MPRC-DQN} or \textbf{KG-A2C}, suggesting NLU techniques might help these agents with RL challenges other than partial observability. Finally, we also show in {Appendix}~\ref{app:gt_room} the performance of replacing our state approximation with the groundtruth room IDs (GT-Room), where our agent achieves on-par or better results on all the games. This confirms that our state approximation not only effectively identifies player locations by \textcolor{black}{Eq. \ref{eq:po1}}, but also brings richer state information thanks to \textcolor{black}{Eq. \ref{eq:po2}}.


\begin{figure}
    \centering
    \includegraphics[width=.45\textwidth]{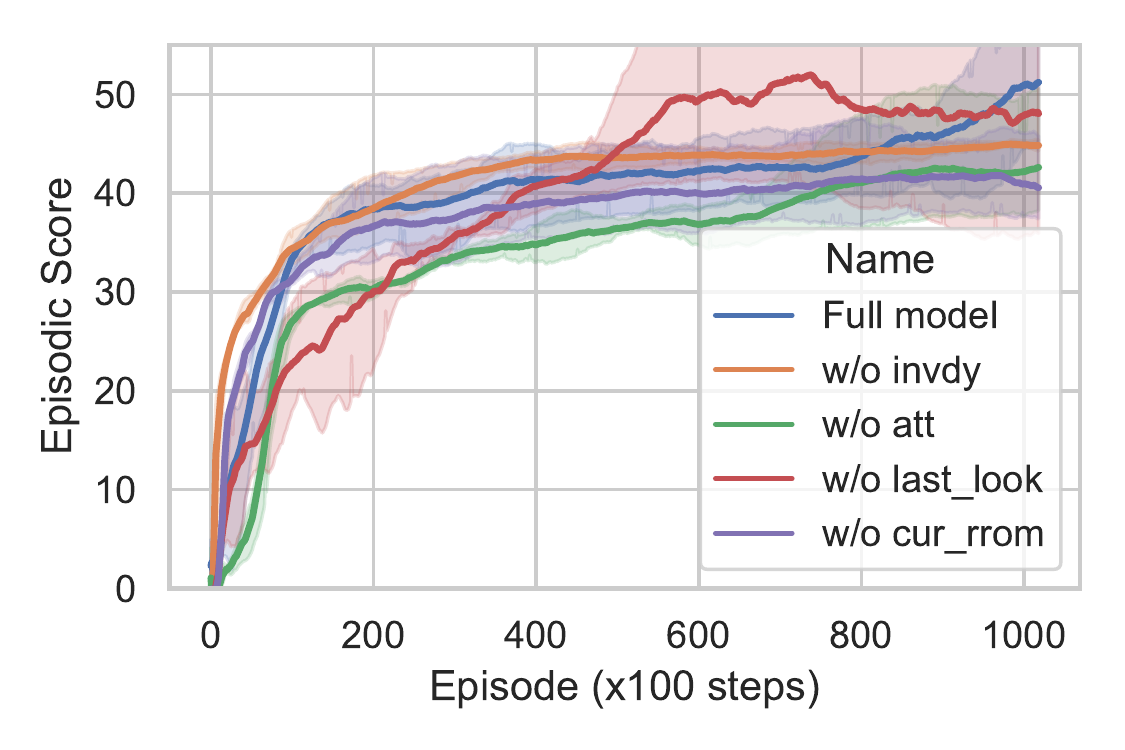}
        \vspace{-5mm}
    \caption{Ablation results on Zork I. }
      \vspace{-6mm}
    \label{fig:ablation}
\end{figure}

\subsection{Ablation Studies}
\label{ssec:ablation_texts}
In light of the good scores of \modelnamens, are distinguishing states and memorizing trajectories all it takes to solve a text game? To answer this question, we conduct ablation experiments to remove the text encoder (i.e.\,all GRUs) in our agent as well as the GT-State version (\textbf{- Text Enc}). Intuitively, this renders the text game into a large, deterministic MDP (instead of POMDP), where even very close states (e.g.\,same except a window is ajar or open) have completely different representations. 

The result in Table~\ref{tab:gt} shows a huge performance drop for \modelname and its GT-State version with text encoders removed --- in other words, learning the text game as a tabular MDP without language semantics could lead to a much deteriorated sample complexity, even when partial observability is solved. 
To explain why DRRN with GT-State hash is much worse than DRRN with observation hash proposed in \citet{yao2021reading}, note that \textcolor{black}{Eq. \ref{eq:sr_hash}} still leverages the compositional structure of $(o_t, i_t, l_t)$, e.g.\,two states with the same $i_t$ still share part of the state representation. Such a result helps confirm the importance of language for the RL challenge of \textbf{large observation and actions spaces}: semantics-preserving function approximation (e.g.\,RNN instead of hash) could be key to interpolation (smooth value estimation for similar states) as well as extrapolation (efficient exploration based on language and commonsense priors). 

Finally, we ablate individual components of \modelname on Zork I. Figure~\ref{fig:ablation} shows that removing the language-learning auxiliary task of inverse dynamics (\textbf{w/o invdy}) or the language attention (\textbf{w/o att}) leads to worse scores, reconfirming that semantic language representations are vital for \modelnamens's success. On the other hand, removing the current whereabouts (\textbf{w/o cur\_room}) leads to much worse performance than removing location descriptions across the map (\textbf{w/o last\_look}), suggesting location identification \textcolor{black}{Eq. \ref{eq:po1}} might be more important for solving partial observability.



\begin{table}
\centering
\small
\resizebox{0.48\textwidth}{!}{
\begin{tabular}{lcccc|cccc}
\toprule
\multirow{3}{*}{\textbf{Game}} & \multicolumn{4}{c}{\bf Ours} & \multicolumn{4}{c}{\bf Ours w/ GT-State}  \\
& \multicolumn{2}{c}{\textbf{Full Model}} & \multicolumn{2}{c}{\textbf{- Text Enc.}} & \multicolumn{2}{c}{\textbf{Full Model}} & \multicolumn{2}{c}{\textbf{- Text Enc.}}  \\
&\bf Avg & \bf Max&\bf Avg & \bf Max&\bf Avg & \bf Max&\bf Avg & \bf Max\\
\midrule
zork1 & 51.2&107 & 4.13&36.3 & 53.6&111 & 6.25&39.3  \\
zork3 & 1.33&5 & 0.85&3 & 1.50&4.7 &1.02&4  \\
pentari & 44.4&60 & 20.3&45 & 46.1&60 & 20&45 \\ 
detective & 288.8&313.3 & 281.3&313.3 & 289.9&310 & 280&290 \\
ludicorp & 16.7&23 & 10.15&22 & 15.9&23 & 9.5&21  \\
inhumane & 25.7&56.7 & 1.9&23 & 24.1&60 & 1.1&20  \\
\midrule
Avg. Norm & .36&.59 & .21&.40 & .37&.59 & .22&.42 \\
\bottomrule
\end{tabular}}
    \vspace{-2mm}

\caption{The results of replacing our state representations with groundtruth state IDs (GT-State Full Model), as well as removing the text encoder (- Text Enc).}
    \vspace{-5mm}

\label{tab:gt}
\end{table}

\section{Discussion}
\label{sec:discussion}



We propose a simple approach to deal with partial observability in text games, which could serve as a competitive baseline for future research, and also inspire similar investigations for other RL challenges to test the limits of memorization and necessity of NLU in different dimensions, which would in turn help identify flaws of current setups and propose better ones. We also hope our idea of best combing semantic and non-semantic language representations could be useful for building next-generation text game agents, as well as for other language applications with memorization needs like closed-domain QA or goal-oriented dialog.


\section*{Limitations}

Our approach to retrieve the global state focuses on different locations.
The simplicity of our method can help prove the value and importance of involving non-semantic representations in text-based games.
Also our hash-based non-semantic representation hide the difference in global state retrieval methods,
as long as they can successfully distinguish different states.
However, we acknowledge that more detailed designs is needed in order to generalize our method to other TBGs.

Another limitation is that our method focuses on text fictions,
a specific type of text-based games.
Most games of this type have lots of locations to explore.
As a result,
our location-based approach can successfully distinguish different states.
Although the direct usage of our approach is limited,
we believe the innovation of combining semantic and non-semantic representations
is helpful in other text-based games and NLP tasks.

\bibliography{anthology,custom}
\bibliographystyle{acl_natbib}

\newpage

\appendix



\section{Algorithm of Our Approximated State Representation Construction}
\label{app:algorithm}

\begin{algorithm}
\caption{Infer the current location with nearby room names using depth-first search with limited depth. This can help distinguish different rooms with the same name in most cases. We use $depth = 1$ in our runs.}
\begin{algorithmic}[1]
\Function{Locate}{$env, depth$}
 \State $state \gets$ current state of $env$
 \State $room \gets$ current room name in $env$
 \State $nearby \gets []$
 \ForAll{$direction$}
  \State step $env$ with action $direction$
  \If{$env$ changed in last action}
    \If{$depth>1$}
     \State $d \gets $ \Call{Locate}{$env, depth - 1$}
    \Else
     \State $d \gets$ current room name in $env$
    \EndIf
    \State append $(direction, d)$ to $nearby$
    \State set $env$ with $state$
  \EndIf
 \EndFor
 \State \Return $nearby$
\EndFunction
\end{algorithmic}
\end{algorithm}

\begin{algorithm}
\caption{Maintain the state approximation with the descriptions the last time we visit each room. The result is then hashed to serve as the \textit{state hash} in our model. For runs provided with grounded room ID, we replace line $3$ with the ground truth.}
\begin{algorithmic}[1]
\State $state \gets \{\}$
\Function{UpdateAndGetState}{$env$}
 \State $room \gets$ \Call{Locate}{$env, depth$}
 \State $look \gets$ \textit{look} of the current state in $env$
 \State $state[room] \gets look$
 \State \Return{$state$}
\EndFunction
\end{algorithmic}
\end{algorithm}

\begin{table*}
\centering
\resizebox{0.5\textwidth}{!}{
\begin{tabular}{lcc|cc|c}
\toprule
\multirow{2}{*}{\textbf{Game}} & \multicolumn{2}{c}{\bf Ours} & \multicolumn{2}{c}{\bf Ours w/ GT-State} & {\textbf{Ours w/}} \\
& {\textbf{Full Model}} & {\textbf{- Text Enc.}} & {\textbf{Full Model}} & {\textbf{- Text Enc.}} & {\textbf{GT-Room}} \\
\midrule
zork1 & 51.2/107 & 4.13/36.3 & 53.6/111 & 6.25/39.3 & 52.0/110 \\
zork3 & 1.33/5 & 0.85/3 & 1.50/4.7 &1.02/4 &1.31/5 \\
pentari & 44.4/60 & 20.3/45 & 46.1/60 & 20/45 & 44.8/58 \\
detective & 288.8/313.3 & 281.3/313.3 & 289.9/310 & 280/290 & 289.6/300 \\
ludicorp & 16.7/23 & 10.15/22 & 15.9/23 & 9.5/21 & 16.9/23 \\
inhumane & 25.7/56.7 & 1.9/23 & 24.1/60 & 1.1/20 & 25.7/56.7 \\
\midrule
Avg. Norm & .36/.59 & .21/.40 & .37/.59 & .22/.42 & .36/.58 \\
\bottomrule
\end{tabular}}
\caption{The results of replacing our state representations with groundtruth state IDs (GT-State Full Model), as well as removing the text encoder (- Text Enc).}
\label{tab:gt2}
\end{table*}

\section{More Details of Our Model and Implementation}
\label{app:detail}
We follow the hyperparameters from \citet{yao2021reading}.
For the state approximation part, we use the builtin \textsc{Hash} function in Python. We train our model for $10^5$ steps, which takes about $40$ hours on a TITAN X or Geforce GTX 1080.

We use the latest Jericho version 3.1.0. Due to a bug in Zork I, we add a timeout in the library to filter out valid actions causing the emulator to hang.

\subsection{Details of Our BiDAF Observation Encoder}
\label{app:bidaf}

In DRRN, the GRU takes the responsibility of both memorizing the high-scoring trajectories, and generalizing to unseen observations.
In our method, the memorization power can be provided with our hash codes of local graphs, with stronger ability to distinguish states.
We thus hope to encourage the generalization strength of neural network; and propose the attentive extension of observation embedding.

Our key idea bases on the insight that the Q-value in DRRN is computed by matching the textual observations to a textual action.
Since the observations are usually significantly longer than the actions, the effect of an action can usually be determined by its interaction with a local context in the observation.
This can be naturally modeled with the attention mechanism. Specifically, we apply the BiDAF~\cite{seo2016bidirectional} to match each observation component to the action.

The BiDAF takes the observation and action embeddings; and outputs an action-attended observation embedding. 
We denote the GRU embeddings for observation word $i$ and action word $j$ as $\bm o_i$ and $\bm a_{j}$. 
The attention score from an observation word to an action word
is thus $\alpha_{ij}{=}\exp(a_{ij})/\sum_{j}\exp(a_{ij})$, where $a_{ij}{=} \bm{o_{i}}^{T} \bm{a_{j}}$.
We then compute the ``action-to-observation'' summary vector for the $i$-th observation word as  $\bm{c_{i}}{=}\sum_{j} \alpha_{ij}\bm{a_{j}}$. 
We concatenate and project the output vectors as $\big[\bm{o_{i}},\allowbreak \bm{c_{i}},\allowbreak \bm{o_{i}}\odot \bm{c_{i}},\allowbreak \vert \bm{o}_i - \bm{c_{i}} \vert \big]$, followed by a linear layer with leaky ReLU activation units.
We apply the aforementioned steps to the inventory $i$ and location appearance $l$, too. Finally, we have
\begin{equation}
\small
\begin{aligned}
    sr_{\text{LoG}} = [&\text{BiDAF}(\bm o, \bm a), \text{BiDAF}(\bm i, \bm a), \text{BiDAF}(\bm l, \bm a),\\
    & \mathrm{H}(po_1(c))), \mathrm{H}(po_2(po_1(c))]
\end{aligned}
\end{equation}




\section{Additional Experiments with Oracle State Information}
\label{app:gt_room}

We investigate of performance of replacing our state approximation with the groundtruth room IDs (GT-Room). 

\textcolor{black}{
Specifically, instead of using our state representation or the groundtruth state ID, we use the groundtruth room ID the player is located in. 
The room ID has a much lower dimension compared to the state ID; and is a simple yet strong indicator for game playing. However, it is only a shortcut for some game states. For fundamentally dealing with partial observability, the state representation should be able to capture more information beyond room IDs.} 

\textcolor{black}{
The experiment here is designed to verify whether our proposed representation can learn more than room IDs.
Intuitively, if a model only learns to mimic room IDs, its performance will be strictly below the oracle GT-Room results. 
While our results in Table 3 show that our model is on par in terms of average measure and is even higher in terms of the maximum measure across all the games. 
This confirms that our state approximation not only identifies the true location of the player, but also brings richer state information.}

\textcolor{black}{
Together with the GT-State results, we proved the advantage of our proposed state representation, as it is on par or better compared to both the sufficient but sparse state-ID oracle and the less informative but smoothed room-ID oracle.}

\section{Zork I Trajectory of \modelnamens}
\textcolor{black}{
Obs0: Copyright (c) 1981, 1982, 1983 Infocom, Inc. All rights reserved. ZORK is a registered trademark of Infocom, Inc. Revision 88 / Serial number 840726  West of House You are standing in an open field west of a white house, with a boarded front door. There is a small mailbox here. Inv: you are empty handed. Desc: west of house you are standing in an open field west of a white house, with a boarded front door. there is a small mailbox here.\\
Action0: north, Score 0\\
\\
Obs1: North of House You are facing the north side of a white house. There is no door here, and all the windows are boarded up. To the north a narrow path winds through the trees. Inv: you are empty handed. Desc: north of house you are facing the north side of a white house. there is no door here, and all the windows are boarded up. to the north a narrow path winds through the trees.\\
Action1: east, Score 0\\
\\
Obs2: Behind House You are behind the white house. A path leads into the forest to the east. In one corner of the house there is a small window which is slightly ajar. Inv: you are empty handed. Desc: behind house you are behind the white house. a path leads into the forest to the east. in one corner of the house there is a small window which is slightly ajar.\\
Action2: open window, Score 0\\
\\
Obs3: With great effort, you open the window far enough to allow entry. Inv: you are empty handed. Desc: behind house you are behind the white house. a path leads into the forest to the east. in one corner of the house there is a small window which is open.\\
Action3: west, Score 10\\
\\
Obs4: Kitchen You are in the kitchen of the white house. A table seems to have been used recently for the preparation of food. A passage leads to the west and a dark staircase can be seen leading upward. A dark chimney leads down and to the east is a small window which is open. On the table is an elongated brown sack, smelling of hot peppers. A bottle is sitting on the table. The glass bottle contains:   A quantity of water Inv: you are empty handed. Desc: kitchen you are in the kitchen of the white house. a table seems to have been used recently for the preparation of food. a passage leads to the west and a dark staircase can be seen leading upward. a dark chimney leads down and to the east is a small window which is open. on the table is an elongated brown sack, smelling of hot peppers. a bottle is sitting on the table. the glass bottle contains:   a quantity of water\\
Action4: west, Score 10\\
\\
Obs5: Living Room You are in the living room. There is a doorway to the east, a wooden door with strange gothic lettering to the west, which appears to be nailed shut, a trophy case, and a large oriental rug in the center of the room. Above the trophy case hangs an elvish sword of great antiquity. A battery powered brass lantern is on the trophy case. Inv: you are empty handed. Desc: living room you are in the living room. there is a doorway to the east, a wooden door with strange gothic lettering to the west, which appears to be nailed shut, a trophy case, and a large oriental rug in the center of the room. above the trophy case hangs an elvish sword of great antiquity. a battery powered brass lantern is on the trophy case.\\
Action5: push rug, Score 10\\
\\
Obs6: With a great effort, the rug is moved to one side of the room, revealing the dusty cover of a closed trap door. Inv: you are empty handed. Desc: living room you are in the living room. there is a doorway to the east, a wooden door with strange gothic lettering to the west, which appears to be nailed shut, a trophy case, and a closed trap door at your feet. above the trophy case hangs an elvish sword of great antiquity. a battery powered brass lantern is on the trophy case.\\
Action6: take all, Score 10\\
\\
Obs7: trophy case: The trophy case is securely fastened to the wall. sword: Taken. brass lantern: Taken. carpet: The rug is extremely heavy and cannot be carried. Inv: you are carrying:   a brass lantern   a sword Desc: living room you are in the living room. there is a doorway to the east, a wooden door with strange gothic lettering to the west, which appears to be nailed shut, a trophy case, and a closed trap door at your feet.\\
Action7: open trap, Score 10\\
\\
Obs8: The door reluctantly opens to reveal a rickety staircase descending into darkness. Inv: you are carrying:   a brass lantern   a sword Desc: living room you are in the living room. there is a doorway to the east, a wooden door with strange gothic lettering to the west, which appears to be nailed shut, a trophy case, and a rug lying beside an open trap door.\\
Action8: open case, Score 10\\
\\
Obs9: Opened. Inv: you are carrying:   a brass lantern   a sword Desc: living room you are in the living room. there is a doorway to the east, a wooden door with strange gothic lettering to the west, which appears to be nailed shut, a trophy case, and a rug lying beside an open trap door.\\
Action9: down, Score 35\\
\\
Obs10: You have moved into a dark place. The trap door crashes shut, and you hear someone barring it.  It is pitch black. You are likely to be eaten by a grue. Your sword is glowing with a faint blue glow. Inv: you are carrying:   a brass lantern   a sword Desc: it is pitch black. you are likely to be eaten by a grue.\\
Action10: turn on lantern, Score 35\\
\\
Obs11: The brass lantern is now on.  Cellar You are in a dark and damp cellar with a narrow passageway leading north, and a crawlway to the south. On the west is the bottom of a steep metal ramp which is unclimbable. Inv: you are carrying:   a brass lantern (providing light)   a sword Desc: cellar you are in a dark and damp cellar with a narrow passageway leading north, and a crawlway to the south. on the west is the bottom of a steep metal ramp which is unclimbable.\\
Action11: north, Score 35\\
\\
Obs12: The Troll Room This is a small room with passages to the east and south and a forbidding hole leading west. Bloodstains and deep scratches (perhaps made by an axe) mar the walls. A nasty looking troll, brandishing a bloody axe, blocks all passages out of the room. Your sword has begun to glow very brightly. Inv: you are carrying:   a brass lantern (providing light)   a sword the axe crashes against the rock, throwing sparks! Desc: the troll room this is a small room with passages to the east and south and a forbidding hole leading west. bloodstains and deep scratches (perhaps made by an axe) mar the walls. a nasty looking troll, brandishing a bloody axe, blocks all passages out of the room. the axe crashes against the rock, throwing sparks!\\
Action12: hit troll with sword, Score 35\\
\\
Obs13: The haft of your sword knocks out the troll. Inv: you are carrying:   a brass lantern (providing light)   a sword Desc: the troll room this is a small room with passages to the east and south and a forbidding hole leading west. bloodstains and deep scratches (perhaps made by an axe) mar the walls. there is a bloody axe here. an unconscious troll is sprawled on the floor. all passages out of the room are open.\\
Action13: west, Score 35\\
\\
Obs14: Maze This is part of a maze of twisty little passages, all alike. Your sword is glowing with a faint blue glow. Inv: you are carrying:   a brass lantern (providing light)   a sword Desc: maze this is part of a maze of twisty little passages, all alike.\\
Action14: west, Score 35\\
\\
Obs15: Maze This is part of a maze of twisty little passages, all alike. Your sword is no longer glowing. Inv: you are carrying:   a brass lantern (providing light)   a sword Desc: maze this is part of a maze of twisty little passages, all alike.\\
Action15: west, Score 35\\
\\
Obs16: Maze This is part of a maze of twisty little passages, all alike. Inv: you are carrying:   a brass lantern (providing light)   a sword Desc: maze this is part of a maze of twisty little passages, all alike.\\
Action16: up, Score 35\\
\\
Obs17: Maze This is part of a maze of twisty little passages, all alike. A skeleton, probably the remains of a luckless adventurer, lies here. Beside the skeleton is a rusty knife. The deceased adventurer's useless lantern is here. There is a skeleton key here. An old leather bag, bulging with coins, is here. Inv: you are carrying:   a brass lantern (providing light)   a sword Desc: maze this is part of a maze of twisty little passages, all alike. a skeleton, probably the remains of a luckless adventurer, lies here. beside the skeleton is a rusty knife. the deceased adventurer's useless lantern is here. there is a skeleton key here. an old leather bag, bulging with coins, is here.\\
Action17: put down sword, Score 35\\
\\
Obs18: Dropped. Inv: you are carrying:   a brass lantern (providing light) Desc: maze this is part of a maze of twisty little passages, all alike. a skeleton, probably the remains of a luckless adventurer, lies here. beside the skeleton is a rusty knife. the deceased adventurer's useless lantern is here. there is a sword here. there is a skeleton key here. an old leather bag, bulging with coins, is here.\\
Action18: take coins, Score 45\\
\\
Obs19: Taken. Inv: you are carrying:   a leather bag of coins   a brass lantern (providing light) Desc: maze this is part of a maze of twisty little passages, all alike. a skeleton, probably the remains of a luckless adventurer, lies here. beside the skeleton is a rusty knife. the deceased adventurer's useless lantern is here. there is a sword here. there is a skeleton key here.\\
Action19: southwest, Score 45\\
\\
Obs20: Maze This is part of a maze of twisty little passages, all alike. Inv: you are carrying:   a leather bag of coins   a brass lantern (providing light) Desc: maze this is part of a maze of twisty little passages, all alike.\\
Action20: put down coins, Score 45\\
\\
Obs21: Dropped. Inv: you are carrying:   a brass lantern (providing light) Desc: maze this is part of a maze of twisty little passages, all alike. an old leather bag, bulging with coins, is here.\\
Action21: take coins, Score 45\\
\\
Obs22: Taken. Inv: you are carrying:   a leather bag of coins   a brass lantern (providing light) Desc: maze this is part of a maze of twisty little passages, all alike.\\
Action22: east, Score 45\\
\\
Obs23: Maze This is part of a maze of twisty little passages, all alike. Inv: you are carrying:   a leather bag of coins   a brass lantern (providing light) Desc: maze this is part of a maze of twisty little passages, all alike.\\
Action23: put down coins, Score 45\\
\\
Obs24: Dropped. Inv: you are carrying:   a brass lantern (providing light) Desc: maze this is part of a maze of twisty little passages, all alike. an old leather bag, bulging with coins, is here.\\
Action24: up, Score 45\\
\\
Obs25: Maze This is part of a maze of twisty little passages, all alike. Inv: you are carrying:   a brass lantern (providing light) Desc: maze this is part of a maze of twisty little passages, all alike.\\
Action25: west, Score 45\\
\\
Obs26: Maze This is part of a maze of twisty little passages, all alike. Inv: you are carrying:   a brass lantern (providing light) Desc: maze this is part of a maze of twisty little passages, all alike.\\
Action26: southeast, Score 45\\
\\
Obs27: Cyclops Room This room has an exit on the northwest, and a staircase leading up. A cyclops, who looks prepared to eat horses (much less mere adventurers), blocks the staircase. From his state of health, and the bloodstains on the walls, you gather that he is not very friendly, though he likes people. Inv: you are carrying:   a brass lantern (providing light) Desc: cyclops room this room has an exit on the northwest, and a staircase leading up. a cyclops, who looks prepared to eat horses (much less mere adventurers), blocks the staircase. from his state of health, and the bloodstains on the walls, you gather that he is not very friendly, though he likes people.\\
Action27: odysse, Score 45\\
\\
Obs28: The cyclops, hearing the name of his father's deadly nemesis, flees the room by knocking down the wall on the east of the room. Inv: you are carrying:   a brass lantern (providing light) Desc: cyclops room this room has an exit on the northwest, and a staircase leading up. the east wall, previously solid, now has a cyclops sized opening in it.\\
Action28: up, Score 70\\
\\
Obs29: You hear a scream of anguish as you violate the robber's hideaway. Using passages unknown to you, he rushes to its defense. The thief gestures mysteriously, and the treasures in the room suddenly vanish.  Treasure Room This is a large room, whose east wall is solid granite. A number of discarded bags, which crumble at your touch, are scattered about on the floor. There is an exit down a staircase. There is a suspicious looking individual, holding a large bag, leaning against one wall. He is armed with a deadly stiletto. There is a silver chalice, intricately engraved, here. The thief draws blood, raking his stiletto across your arm. Inv: you are carrying:   a brass lantern (providing light) you dodge as the thief comes in low. Desc: treasure room this is a large room, whose east wall is solid granite. a number of discarded bags, which crumble at your touch, are scattered about on the floor. there is an exit down a staircase. there is a suspicious looking individual, holding a large bag, leaning against one wall. he is armed with a deadly stiletto. there is a silver chalice, intricately engraved, here. you dodge as the thief comes in low.\\
Action29: temple, Score 70\\
\\
Obs30: Temple This is the north end of a large temple. On the east wall is an ancient inscription, probably a prayer in a long forgotten language. Below the prayer is a staircase leading down. The west wall is solid granite. The exit to the north end of the room is through huge marble pillars. There is a brass bell here. Inv: you are carrying:   a brass lantern (providing light) Desc: temple this is the north end of a large temple. on the east wall is an ancient inscription, probably a prayer in a long forgotten language. below the prayer is a staircase leading down. the west wall is solid granite. the exit to the north end of the room is through huge marble pillars. there is a brass bell here.\\
Action30: throw lantern at prayer, Score 70
}


\end{document}